\documentclass{article}

\usepackage{arxiv}
\usepackage{mathtools}
\usepackage[utf8]{inputenc} 
\usepackage[mathcal]{eucal}
\usepackage[T1]{fontenc}    
\usepackage{hyperref}       
\usepackage{url}            
\usepackage{booktabs}       
\usepackage{amsfonts}       
\usepackage{nicefrac}       
\usepackage{microtype}      
\usepackage{lipsum}
\usepackage{fancyhdr}       
\usepackage{graphicx}       
\usepackage[ruled,vlined]{algorithm2e}  
\graphicspath{{media/}}     

\title{SS-MFAR:Semi-supervised Multi-task Facial Affect Recognition
}

\author{
  Dr. Darshan Gera \\
  SSSIHL, Brindavan Campus \\
  Bengaluru, Karnataka, India\\
  \texttt{darshangera@sssihl.edu.in} \\
  \And
  Badveeti Naveen Siva Kumar\\
  SSSIHL, Prasanthi Nilayam Campus \\
  Sri Sathya Sai District, Andhra Pradesh, India \\
  \texttt{bnaveensivakumar@gmail.com} \\
  \AND
 Bobbili Veerendra Raj Kumar\\
  SSSIHL, Prasanthi Nilayam Campus \\
  Sri Sathya Sai District, Andhra Pradesh, India \\
  \texttt{veerendra.rajkumar@gmail.com} \\
  \And
  Dr. S Balasubramanian \\
  SSSIHL, Prasanthi Nilayam Campus \\
  Sri Sathya Sai District, Andhra Pradesh, India\\
  \texttt{sbalasubramanian@sssihl.edu.in} \\
}

\hypersetup{colorlinks=true, urlcolor=cyan,
pdftitle={Multi Task Learning for Affect Recognition at ABAW Competition Using Semi Supervised Learning},
pdfauthor={Darshan Gera, S Balasubramanian, Badveeti Naveen Siva Kumar, Bobbili Veerendra Raj Kumar},
pdfkeywords={Multi Task Learning, Semi Supervised Learning, Facial Expression Recognition, AffWild2, s-aff-wild2},
}

\begin{document}
\maketitle

\begin{abstract}
Automatic affect recognition has applications in many areas such as education, gaming, software development, automotives, medical care, etc. but it is non trivial task to achieve appreciable performance on in-the-wild data sets. In-the-wild data sets though represent real-world scenarios better than synthetic data sets, the former ones suffer from the problem of incomplete labels. Inspired by semi-supervised learning, in this paper, we introduce our submission to the Multi-Task-Learning Challenge at the 4th Affective Behavior Analysis in-the-wild (ABAW) 2022 Competition. The three tasks that are considered in this challenge are valence-arousal(VA) estimation, classification of expressions into 6 basic (anger, disgust, fear, happiness, sadness, surprise), neutral, and the 'other' category and 12 action units(AU) numbered AU-\{1,2,4,6,7,10,12,15,23,24,25,26\}. Our method Semi-supervised Multi-task Facial Affect Recognition titled \textbf{SS-MFAR} uses a deep residual network with task specific classifiers for each of the tasks along with adaptive thresholds for each expression class and semi-supervised learning for the incomplete labels. Source code is available at \url{https://github.com/1980x/ABAW2022DMACS}.
\end{abstract}


\section{Introduction}
Automatic affect recognition is currently an active area of research and has applications in many areas such as education, gaming, software development, auto motives \cite{appsofFER, sym14040687}, medical care, etc.  Many works have dealt with Valence-arousal estimation \cite{oh2021causalpast, VARealtime, oh2021causal, VAmeng2022, deepVA18}, action unit detection \cite{jacob2021AU, shao2019AU, tang2017AU}, expression classification \cite{} tasks individually. \cite{XIAOHUA2019MTL} introduced a framework which uses only static images and Multi-Task-Learning (MTL) to learn categorical representations and use them to estimate dimensional representation, but is limited to AffectNet data set \cite{12}. Aff-wild2 \cite{kollias2018aff, kollias2019expression, kollias2018multi, kollias2020analysing, kollias2021distribution, kollias2021affect, kollias2019deep, kollias2020va,zafeiriou2017aff, kollias2017recognition} is the first dataset with annotations with all three tasks. \cite{kollias2021distribution} study uses MTL for all the three tasks mentioned earlier along with facial attribute detection and face identification as case studies to show that their network FaceBehaviourNet learns all aspects of facial behaviour. It is shown to perform better than the state-of-the-art models of individual tasks. One of major limitation of Affwild2 is that annotations of valence-arousal, AU and expression are not available for all the samples. So, this dataset has incomplete labels for different tasks and further imbalanced for different expression classes. In a recent work, \cite{AdaCM} uses semi-supervised learning with adaptive confidence margin for expression classification task for utilizing unlabelled samples. Adaptive confidence margin is used to deal with inter and intra class difference in predicted probabilities for each expression class. In this work, we developed a Multi-task Facial Affect Recognition (\textbf{MFAR}) in which a ResNet-18 \cite{he2016deep} network pre-trained on MS-Celeb-1M \cite{msceleb} weights is used, along with task specific classifiers for the three tasks. To handle the class imbalance in the dataset, we introduced re-weighting. From the s-Aff-Wild2 \cite{abaw2022} dataset, we observed that out of $142383$ a total of $51737$ images have invalid expression annotations. Motivated from \cite{AdaCM}, we added semi-supervised learning (\textbf{SS-MFAR}) to label the images with invalid expression annotations.  

\section{Method}
In this section, we present our solution to the MTL Challenge at the 4th Affective Behavior Analysis in-the-wild (ABAW) Competition. 
\subsection{MFAR : Multi-Task Facial Affect Recognition} \label{sec:mfar}
\begin{figure*}[ht!] 
    \centering
    \includegraphics[width=1.0\textwidth]{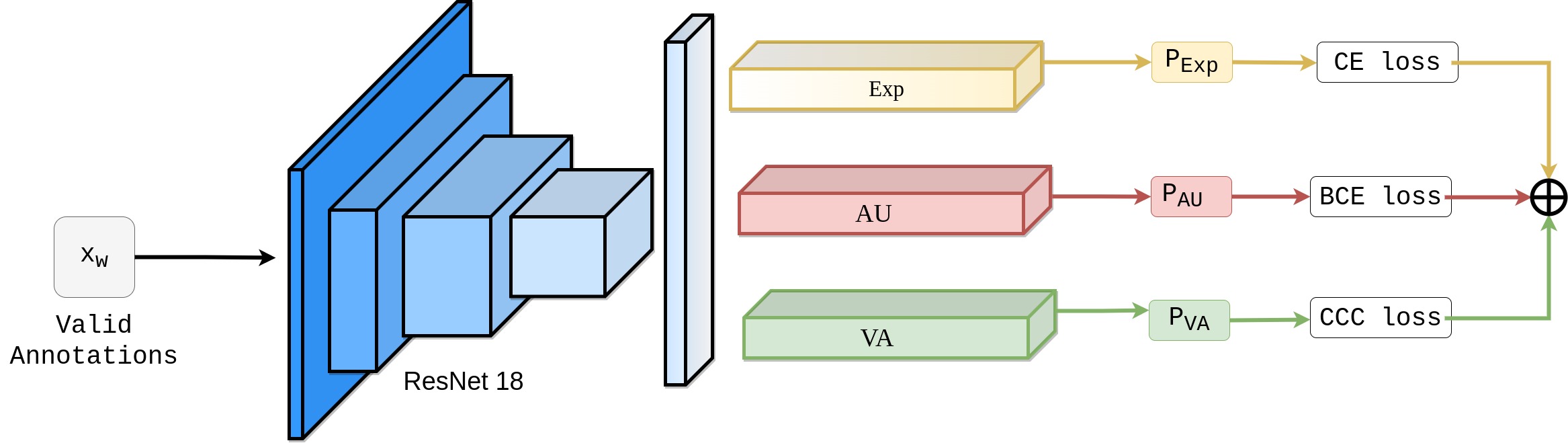}
    \caption{\textbf{MFAR} architecture is shown in this figure. Weak Augmentations $(x_w)$ of input images with valid annotations are fed to the network and the outputs from each of the tasks are obtained. Losses for each task are added to get the overall loss function$^{\ref{Overallmfarloss}}$, that MFAR minimizes. CE is cross entropy$^{\ref{lossCE}}$, BCE is binary cross entropy loss$^{\ref{lossBCE}}$, CCC loss is Concordance Correlation Coefficient based loss$^{\ref{lossCCC}}$.}
    \label{fig:mfar}
\end{figure*}

\textbf{MFAR} architecture is shown in the Figure \ref{fig:mfar}. Weak Augmentations ($x_w$) of input images which have valid annotations are fed to the network and the outputs from each of the tasks are obtained. In s-Aff-Wild2 dataset which is the training set used valid annotations for 
$i)$ expression is any integer value in \{0,1,2,\dots,7\}, 
$ii)$ action unit annotations is either 0 or 1 and 
$iii)$ valence-arousal is any value in the range [-1, 1], but there are images that have expression and action unit labels annotated with $-1$, and valence-arousal values annotated as $-5$ which are treated as invalid annotations. Outputs obtained from each of the task specific classifier is used to find task related loss and then combined to give the overall loss$^{\ref{Overallmfarloss}}$ the \textbf{MFAR} network minimizes. Losses used for each task are \textit{Cross Entropy}, \textit{Binary Cross Entropy}, \textit{Concordance Correlation Coefficient} based loss for expression classification, action unit detection and valence-arousal estimation tasks respectively. Re-scaling weight for each class was calculated using Eq.\ref{eq:expRW} and used to counter the imbalance class problem.
If $N_{Exp}$ is the total number of valid expression samples and $n_{Exp}[i]$ is the number of valid samples for each expression class, then weights for a class $i$ is denoted by $W_{Exp}[i]$ and is defined as 
\begin{equation} \label{eq:expRW}
     W_{Exp}[i] = {\frac{N_{Exp}}{n_{Exp}[i]}}
\end{equation}

Positive weight, which is the ratio of negative samples and positive samples for each action unit as in Eq.\ref{eq:auRW} is used in binary cross entropy loss to counter the imbalance in the classes that each action unit can take. If $N_{AU}^P[i]$ is the number of positive samples and $N_{AU}^N[i]$ is the number of negative samples for each action unit, then positive weight for action unit i is denoted by $W_{AU}[i]$ and is defined as 
\begin{equation} \label{eq:auRW}
     W_{AU}[i] = {\frac {N_{AU}^N[i]} {N_{AU}^P[i]} }  
\end{equation}
The overall loss is the sum of losses of each task. \textbf{MFAR} network minimizes the overall loss function$^{\ref{Overallmfarloss}}$.

\begin{equation} \label{Overallmfarloss}
    {L}_{MFAR} = {L}_{VA} + {L}_{AU} + {L}_{Exp}
\end{equation}

\subsection{SS-MFAR : Semi-Supervised Multi-Task Facial Affect Recognition}

\begin{figure*}[ht!] 
    \centering
    \includegraphics[width=1.0\textwidth]{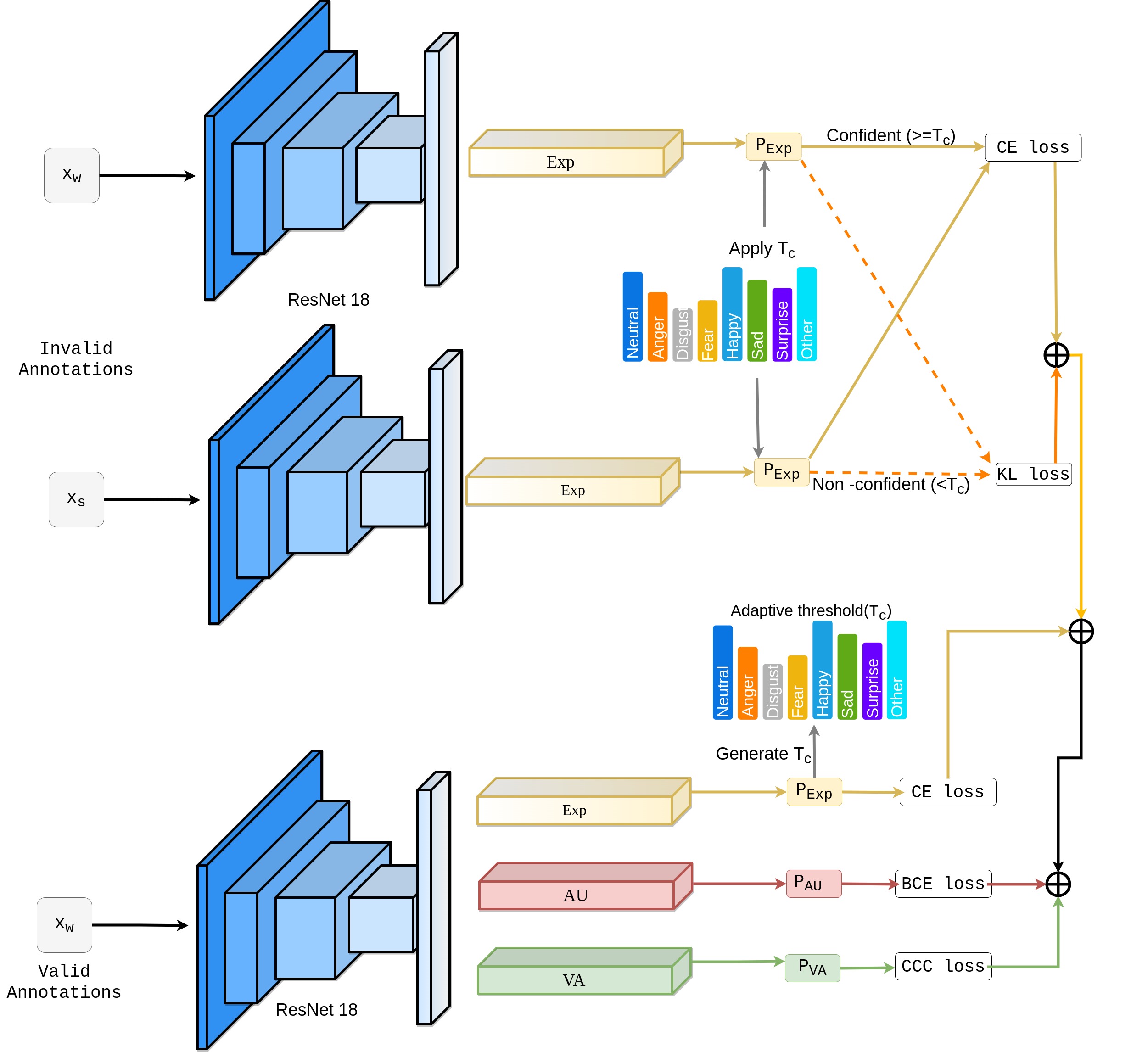}
    \caption{\textbf{SS-MFAR} architecture of our proposed solution. Along with obtaining predictions on the input like \textbf{MFAR}, adaptive threshold for each expression class is learnt. This adaptive threshold is used to mark the  predictions on input images as confident and non-confident ones. CE is cross entropy loss$^{\ref{lossCE}}$ and is used on the confident predictions of weak and strong augmentations, KL is symmetric kl-divergence$^{\ref{lossKL}}$ between the probability distributions of non-confident weak and strong predictions and the remaining losses are as in \textbf{MFAR}.}
    \label{fig:ssmfar}
\end{figure*}

\textbf{SS-MFAR} architecture is shown in the Figure \ref{fig:ssmfar}. \textbf{SS-MFAR} is our proposed solution to the MTL challenge. \textbf{SS-MFAR} adds \emph{semi-supervised learning} to \textbf{MFAR} on the images that have \emph{invalid expression annotations}. The images that have invalid expression annotations i.e., have expression label as -1 are treated as unlabelled data and therefore inspired from \cite{AdaCM} we use semi-supervised learning to predict their labels thereby utilizing more images than \textbf{MFAR}$^{\ref{sec:mfar}}$ along with adaptive threshold to deal with inter and intra class differences in the prediction probabilities that arise due to input samples that are easy and hard within each expression class and among the expression classes. The overall loss function \textbf{SS-MFAR} minimizes is the sum of losses of each task as shown in Eq.\ref{overallssmfarloss}. Losses\ref{Overallmfarloss} with respect to AU detection task and VA estimation task are same as in \textbf{MFAR}. For loss with respect to expression task, we use weighted combination of CE on labelled samples, CE on confident predictions of $x_s$ and $x_w$, KL on non-confident predictions of $x_w$ and $x_s$, where CE is cross entropy loss with class-specific re-scaling weights$^{\ref{eq:expRW}}$, KL is symmetric KL divergence as given in Eq.\ref{lossKL}.
 
\subsection{Problem formulation}
Let $D = {\{(x_i, \tilde{y}_i)\}}^N_{i=1}$ be the  dataset of N samples. Here $x_{i}$ is $i^{th}$ image where $\tilde{y}_i$ represents expression class ${y}^{Exp}_i)$, action unit annotations ${y}^{AU}_i)$ and valence arousal annotations ${y}^{VA}_i)$ of $i^{th}$ image. The backbone network is parameterized by $\theta$ ( ResNet-18 pre-trained on MS-Celeb-1M is used as the backbone). We denote $x_w$ as weak augmented image and $x_s$ as strong augmented image, $P_{Exp}$ represent probability distribution predicted by Expression classifier, $P_{AU}$ represent probability of action unit predicted by AU classifier and $P_{VA}$  of valence arousal predicted by VA classifier. Weak augmentations include standard cropping and horizontal flipping of the input image. Strong augmentations includes weak augmentations along with Randaugment \cite{Randaug}.

\subsection{Adaptive Threshold}

To tackle inter and intra class differences in prediction probabilities of expression, we use adaptive threshold for each class. Here we generate the threshold based on the predictions probabilities whose prediction label matches the ground truth label and threshold is also a function of epoch number since the discriminative ability of the model increases as epoch number increases.We denote the ground truth labels with $\tilde{y}_i$ for image $x_i$ ,ep as epoch number, $p_i$ denotes prediction probabilities of the image i and  we denote adaptive threshold as $T^c$ where c $\in$ $\{1,2,...,8\}$ defined as:
\begin{equation} \label{thresh}
    T^c = \frac{\beta*({\frac{1}{N^s}}\sum^{N^s}_{i=1}\delta^c_i*p_i)}{1+\gamma^{-ep}} \hspace*{1em},\hspace*{1em} \mathrm{where}
\end{equation}
\begin{equation*} 
    \delta^c_i = \left \{ 
                        \begin{array}{ll}
                            1 & \mathrm{if \hspace*{1em}}  \tilde{y}_i = c, \\ \\
                            0 & \mathrm{otherwise}.
                        \end{array}
    \right.
\end{equation*}
The values  $\beta=0.95$ and $\gamma=e$ are taken from \cite{AdaCM} 
\subsection{Supervision loss}

Supervision loss is computed based on the weak augmented images $x_w$ that have valid annotations for each task. We use Cross Entropy loss for the expression classification, binary cross entropy loss for AU detection and  Concordance Correlation Coefficient loss for VA estimation.

Cross Entropy loss denoted by $L_{CE}((X,Y),\theta)$ used for expression classification is
\begin{equation} \label{lossCE}
L^s_{CE} = (-\sum^8_{c=1} \tilde{y}^{Exp^c}_{i=1}log(p^c(x_i,\theta))) ) 
\end{equation} 
 
Binary Cross Entropy loss denoted by $L_{BCE}((X,Y),\theta)$ used for action unit detection is
\begin{equation} \label{lossBCE}
L_{BCE} = (-\sum^2_{c=1} \tilde{y}^{AU^c}_{i=1}log(p^c(x_i,\theta))) ) 
\end{equation}

Concordance Correlation Coefficient loss denoted by $L_{CCC}((X,Y),\theta)$ used for valence arousal estimation is
\begin{equation} \label{lossCCC}
    L_{CCC} = 1-{\frac{2*s_{xy}}{s^2_{x}+s^2_{x}+(\bar{x}-\bar{y})^2}}
\end{equation}

\subsection{Unsupervised and Consistency Loss}

Unsupervised Loss and Consistency Loss are used to learn from the images that have invalid expression annotations. We send weak augmented and strong augmented images to the expression classifier and based on the threshold generated earlier, we mark all the images that we have taken into confident and non-confident predictions. Unsupervised Loss is  Cross Entropy loss denoted by $L^u_{CE}$ on the confident logits of strong augmentations and labels predicted on weak augmentations.Consistency Loss is symmetric KL loss on the predicted probability distributions of weak and strong augmentations. 

Symmetric KL-loss is defined for (p,q) probability distributions as :
\begin{equation} \label{lossKL}
    L^c_{KL} =  p*log(\frac{p}{q}) + q*log(\frac{q}{p})
\end{equation}
\subsection{Overall Loss}

Overall loss is linear combination of losses from each task, where loss from Expression classification is defined as $ \mathcal{L}_{Exp} = \lambda_1*L^s_{CE} + \lambda_2*L^u_{CE}  + \lambda_3*L^c_{KL} $ where $ \lambda_1 = 0.5, \lambda_2 = 1, \lambda_3 = 0.1 $ taken from \cite{AdaCM} , loss from AU detection is defined as $ \mathcal{L}_{AU} = L_{BCE}$ and loss from VA estimation is defined as $ \mathcal{L}_{VA} = L_{CCC}$.
Overall loss is defined as 
\begin{equation} \label{overallssmfarloss}
    \mathcal{L}_{Overall} = \mathcal{L}_{Exp} + \mathcal{L}_{AU} + \mathcal{L}_{VA}
\end{equation}

\begin{algorithm}[H]
\SetAlgoLined
 \textbf{INPUT:} dataset(D), parameters($\theta$), model(ResNet-18) pretrained on MS-Celeb,$\eta$(learning rate)\\
  1.for epoch = 1,2,3,...,$epoch_{max}$\\
  2.\hspace*{1em} for i =1,2,3,...,$N_B$\\
  3.\hspace*{1em}\hspace*{1em}Obtain logits from model which has valid annotations for each task from $x_w$\\
  4.\hspace*{1em}\hspace*{1em}Obtain Adaptive threshold$(T_c)$ according to Eq. \ref{thresh}\\
  5.\hspace*{1em}\hspace*{1em}Obtain logits for $x_w$ and $x_s$ for invalid annotations of expression class\\
  6.\hspace*{1em}\hspace*{1em}If prediction probability $p_i > T_c$:\\
  7.\hspace*{1em}\hspace*{1em}\hspace*{1em}Mark those images predictions as confident otherwise mark them non confident.\\
  8.\hspace*{1em}\hspace*{1em}Compute overall loss according to Eq. \ref{overallssmfarloss} using Eqs. \ref{lossCE},\ref{lossBCE},\ref{lossCCC},\ref{lossKL}\\
  9.\hspace*{1em}\hspace*{1em}Update $\theta$ using Eq. \ref{overallssmfarloss}
  
 \caption{Multi-Task Facial Affect Recognition Using Semi-Supervised Learning}
 \label{Alg-ssmfar}
\end{algorithm}

\section{Dataset and Implementation Details}

\subsection{Dataset}
s-AffWild2 \cite{abaw2022} database is a static version of Aff-Wild2 database and contains a total of  
220419 images. It divided into training, validation and test sets with 142383, 26877 and 51159 number of images respectively. The following observations were made with respect to training data:
\begin{itemize}
    \item 38465, 39066, 51737 images have invalid valence-arousal, action unit, expression annotations respectively.
    \item for an image if valence has an invalid annotation, then so is arousal.
    \item for an image if one of the action unit's has invalid annotation, then so does all the other.
\end{itemize}
Cropped-aligned images were used for training the network. The dataset contains valence-arousal, expression and action unit annotations. Values of valence-arousal 
 are in the range[-1,1], expression labels span over eight classes namely anger, disgust, fear, happiness, sadness, surprise, neutral, and other. 12 action units were considered, specifically AU-{1,2,4,6,7,10,12,15,23,24,25,26}.

\subsection{Implementation Details}

\textbf{MFAR} model consists of ResNet-18 network loaded with pre-trained weights of MS-celeb-1M, feature maps of the input were taken out from the last third layer of network which were fed through average pool, dropout and flatten layers to obtain features of the image which were then normalized. These normalized features were then passed to each of the task specific network for further processing. The expression classifier, which is a multi-class classifier consists of a linear layer with ReLU activation followed by a linear layer with output dimensions equal to the number of expression classes(8) addressed in this challenge. The action unit classifier consists of 12 binary classifiers which are addressed as part of this challenge. For valence arousal task the normalized features were passed through a fully connected layer with ReLU activation to obtain the output logits.
\textbf{SS-MFAR} model uses the same network as \textbf{MFAR} but gets the predictions of invalid expression annotations and uses these predictions with appropriate losses to better the overall performance on the given task. 
The proposed methods were implemented in PyTorch using GeForce RTX 2080 Ti GPUs with 11GB memory. The backbone network used is ResNet-18 pre-trained on large scale face dataset MS-Celeb-1M. All the cropped-aligned of s-Aff-Wild2, provided by organizers were used after resizing them to 224x224. Batch size is set to 256. Optimizer used is Adam. Learning rate (lr) is initialized as 0.001 for base networks and 0.01 for the classification layer.

\subsection{Evaluation Metrics}
The overall performance score used for MTL challenge consists of : 
\begin{itemize}
    \item sum of the average of Concordance Correlation Coefficient (CCC) of valence and arousal,
    \item average F1 Score of the 8 expression categories (i.e., macro F1 Score)
    \item average F1 Score of the 12 action units (i.e., macro F1 Score)
\end{itemize}
The overall performance for the MTL challenge is given by:
\begin{equation}
\mathcal{P}_{MTL} = \mathcal{P}_{VA} + \mathcal{P}_{AU} + \mathcal{P}_{Exp}
                  =  {\frac{\rho_{a}+\rho_{v}}{2} + \frac{\sum_{Exp}F^{Exp}_{1}}{8} + \frac{\sum_{AU}F^{AU}_{1}}{12}}
\end{equation}
CCC is defined as 
\begin{equation}
\rho_{c} = {\frac{2*s_{xy}}{s^2_{x}+s^2_{x}+(\bar{x}-\bar{y})^2}}
\end{equation}

F1 score is defined as harmonic mean of precision (i.e. Number of positive class images correctly identified out of
positive predicted) and recall (i.e. Number of positive class images correctly identified out of true positive class). It can
be written as:
\begin{equation}
F_{1} = {\frac{2*precision*recall}{precision+recall}}
\end{equation}

\subsection{Experiments}
Along with the models presented until now many other approaches to the given challenge were tried out. We list few of them here and the results obtained in Table \ref{tab:Tab1}.
\begin{itemize} 
    \item In order to deal with the expression class imbalance, instead of using the re-weighting technique on \textbf{MFAR}, re-sampling technique was used. We call this approach \textbf{MFAR-RS}.
    \item To see the importance of consistency loss we ran a model without KL loss \ref{lossKL}. We call this approach  \textbf{SS-MFAR-NO\_KL}.
    \item In order to deal with the expression class imbalance, instead of using the re-weighting technique on \textbf{SS-MFAR}, re-sampling technique was used. We call this approach \textbf{SS-MFAR-RS}.
    \item Instead of using the pre-trained  weights of MS-Celeb-1M, network pre-trained on AffectNet \cite{12} database was used. We call this approach \textbf{SSP-MFAR}. 
    \item Instead of just using unsupervised loss $L^u_{CE}$ on only non-confident expression predictions, unsupervised loss on $x_s$ and $x_w$ of all the images were added to respective task losses. We call this approach \textbf{SSP-MFAR-SA}.
    \item Instead using a backbone network of ResNet-18 we used a backbone network ResNet-50 on the best performing model without any pre-trained weights. We call this approach \textbf{SS-MFAR-50}.
\end{itemize}

\section{Results and discussion}
We report our results on the official validation set of MTL from the ABAW 2022 Challenge \cite{abaw2022} in Table \ref{tab:Tab1} . Our best performance achieves overall score of 1.125 on validation set which is a significant improvement over baseline.

\begin{table}[hbt!]
\centering
    \caption{Performance comparison on s-Aff-Wild2 validation set}
    \begin{tabular}{c|c|c|c|c}
         \hline
         Method & Exp-F1 score & AU-F1 score & VA-Score & Overall \\
         \hline
         \hline
         Baseline \cite{abaw2022} & - & - & -          & \textit{0.30} \\
         MFAR     &  \textit{0.222}            & \textit{0.493}         & \textit{0.328}           & \textit{1.043}\\
         MFAR-RS  & \textit{0.191}    & \textit{0.40} & \textit{0.375} & \textit{0.966}\\
         SS-MFAR  & \textbf{0.235}    & \textbf{0.493} & \textbf{0.397} & \textbf{1.125}\\
         SS-MFAR-RS  & \textit{0.256}    & \textit{0.461} & \textit{0.361} & \textit{1.078}\\
         SS-MFAR-NO\-KL  & \textit{0.205}    & \textit{0.454} & \textit{0.357} & \textit{1.016}\\
         SSP-MFAR  & \textit{0.233}    & \textit{0.497} & \textit{0.378} & \textit{1.108}\\
         SSP-MFAR-SA  & \textit{0.228}    & \textit{0.484} & \textit{0.40} & \textit{1.112}\\
         SS-MFAR-50  & \textit{0.168}    & \textit{0.425} & \textit{0.296} & \textit{0.889}\\
         \hline
    \end{tabular}
    \label{tab:Tab1}
\end{table}

In addition, we present results on validation set of Synthetic Data Challenge using the proposed method in Table \ref{tab:Tab2} for 6 expression classes.
\begin{table}[hbt!]
\centering
    \caption{Performance comparison on validation set of Synthetic Data Challenge}
    \begin{tabular}{c|c}
         \hline
         Method &  Exp-F1 score \\
         \hline
         \hline
         Baseline \cite{abaw2022} & 0.30 \\
         SS-MAFR    & \textbf{0.587} \\
         \hline
    \end{tabular}
    \label{tab:Tab2}
\end{table}

\subsection{Ablation Studies}
In this section we present our studies with multiple approaches taken in order to solve the given problem in this challenge. 
\subsubsection{Influence of Resampling}
We see from Table \ref{tab:Tab1} that resampling technique used in \textbf{MFAR-RS} doesn't boost the performance of the model as much as re-weighting does in \textbf{SS-MFAR}.

\subsubsection{Impact of Consistency loss}
We can see from the Table \ref{tab:Tab1} the ill effect on model performance when consistency loss \ref{lossKL} is not used. SS-MFAR-NO\-KL when compared with \textbf{SS-MFAR}, the overall performance drops by 0.103.

\subsubsection{Effect of pretrained weights}
\textbf{SS-MFAR} used the pre-trained weights of MS-Celeb-1M, similarly \textbf{SSP-MFAR} used the pre-trained weights of AffectNet database and obtained relatable performance. We also ran an experiment using the model of \textbf{SS-MFAR} without pre-trained weights but obtained poor performance, similarly we see SS-MFAR-50 performance is on par with \textbf{SS-MFAR} model.

\subsection{Performance on Test set}
The performance of SS-MFAR on official test set of MTL challenge is presented in Table \ref{tab:Tab5} and on official test set of LSD challenge is presented in Table \ref{tab:Tab4}. Clearly, our model performs significantly better than Baseline for both the challenges. In case of MTL challenge, SS-MFAR is able to perform better than methods like ITCNU, USTC-AC \cite{ustcac}, DL-ISIR. Even though its performance is lower compared to methods like HUST-ANT \cite{hustant}, STAR-2022 \cite{star2022}, CNU-Sclab \cite{cnusclab}, HSE-NN \cite{hsenn} and Situ-RUCAUM3 \cite{situRUCAIM3}, our method is first of its kind to use all the samples for representation learning based on semi-supervised learning and it is computationally efficient. In case of LSD challenge, our method by using a simple residual network is able to outperform methods like USTC-AC \cite{ustcac_lsd} and STAR-2022 \cite{star2022}.

\begin{table}[hb!]
 \centering
    \caption{Performance comparison on s-Aff-Wild2 Test set of MTL challenge (Refer \url{https://ibug.doc.ic.ac.uk/resources/eccv-2023-4th-abaw/} for *)}
    \begin{tabular}{c|c}
         \hline
          Method &   Overall \\
         \hline
         \hline
          Baseline \cite{abaw2022} & 28.00  \\
         
          ITCNU*  & 68.85  \\
         
         USTC-AC \cite{ustcac} & 93.97 \\
         DL\_ISIR*  &   101.87 \\
         HUST-ANT \cite{hustant} & 107.12  \\         
         STAR-2022 \cite{star2022} & 108.55 \\
         CNU-Sclab \cite{cnusclab} & 111.35  \\
         HSE-NN \cite{hsenn}& 112.99  \\
         Situ-RUCAIM3 \cite{situRUCAIM3} & 143.61 \\
          
         \hline
         SS-MFAR & 104.06 \\
         \hline
    \end{tabular}
    \label{tab:Tab5}
\end{table}

\begin{table}[ht!]
 \centering
    \caption{Performance comparison on Test set of LSD challenge}
    \begin{tabular}{c|c}
         \hline
          Method &   Overall \\
         \hline
         \hline
          Baseline \cite{abaw2022} & 30.00  \\                
         USTC-AC \cite{ustcac} & 30.92 \\
         STAR-2022 \cite{star2022} & 32.40 \\
         HUST-ANT \cite{hustant} & 34.83 \\                  
         ICT-VIPL* & 34.83 \\
         PPAA* & 36.51 \\
         HSE-NN \cite{hsenn} & 37.18  \\                  
          
         \hline
         SS-MFAR & 33.64 \\
         \hline
    \end{tabular}
    \label{tab:Tab4}
\end{table}

\section{Conclusions}
In this paper, we presented our proposed Semi-supervised learning based Multi-task Facial Affect Recognition framework (SS-MFAR) for ABAW challenge conducted as a part of ECCV 2022. SS-MFAR used all the samples for learning the features for expression classification, valence-arousal estimation and action unit detection by learning adaptive confidence margin. This adaptive confidence margin was used to select confident samples for supervised learning from different expression classes overcoming inter class difficulty and intra class size imbalance. The non-confident samples were used by minimizing the unsupervised consistency loss between weak and strong augmented view of input image. The experiments demonstrate the superiority of proposed method on s-Aff-Wild2 dataset. 

\section*{Acknowledgments}
We dedicate this work to Our Guru and Guide Bhagawan Sri Sathya Sai Baba, Divine Founder Chancellor of Sri Sathya Sai Institute of Higher Learning, Prasanthi Nilayam, Andhra Pradesh, India. 

\bibliographystyle{unsrt}
\bibliography{references}  

\end{document}